\def\year{2018}\relax
\documentclass[letterpaper]{article} 
\usepackage{aaai18}  
\usepackage{times}  
\usepackage{helvet}  
\usepackage{courier}  
\usepackage{url}  
\usepackage{graphicx}  
\usepackage{booktabs}       
\usepackage{amsfonts}       
\usepackage{nicefrac}       
\usepackage{microtype}      

\usepackage{amsmath}
\usepackage{amssymb}

\usepackage{algorithm}
\usepackage{algorithmic}
\usepackage{appendix}
\usepackage{multirow}
\usepackage{bbm}
\usepackage{caption}
\usepackage{subcaption}
\usepackage{color}

\DeclareMathOperator*{\argmax}{arg\,max}
\DeclareMathOperator*{\argmin}

\newcommand{\namecite}[1]{\citeauthor{#1}~(\citeyear{#1})}

\begin{document}
%

\title{BBQ-Networks: Efficient Exploration in Deep Reinforcement Learning\\ for Task-Oriented Dialogue Systems}

\author{Zachary Lipton$^\star$\quad Xiujun Li$^\dag$\quad Jianfeng Gao$^\dag$\quad Lihong Li$^\ddag$\thanks{This work was done while ZL, LL \& LD were with Microsoft.}\quad Faisal Ahmed$^\dag$\quad Li Deng$^\S$$^*$\\
$^\star$Carnegie Mellon University, Pittsburgh, PA, USA \\
$^\star$Amazon AI, Palo Alto, CA, USA \\  $^\dag$Microsoft Research, Redmond, WA, USA\\
$^\ddag$Google Inc., Kirkland, WA, USA\quad\quad \\
$^\S$Citadel, Seattle, WA, USA\\
{\small \tt $^\star$zlipton@cmu.edu, $^\dag$\{xiul,jfgao,fiahmed\}@microsoft.com} \\
{\small \tt $^\ddag$lihongli.cs@gmail.com, $^\S$l.deng@ieee.org}
}
 
\maketitle

\begin{abstract} 
We present a new algorithm that significantly improves the efficiency of exploration for deep Q-learning agents in dialogue systems. Our agents explore via Thompson sampling, drawing Monte Carlo samples from a \emph{Bayes-by-Backprop} neural network. Our algorithm learns much faster than common exploration strategies such as $\epsilon$-greedy, Boltzmann, bootstrapping, and intrinsic-reward-based ones. Additionally, we show that spiking the replay buffer with experiences from just a few successful episodes can make Q-learning feasible when it might otherwise fail.
\end{abstract} 

\section{Introduction}
\label{sec:introduction}
Increasingly, we interact with computers 
via natural-language dialogue interfaces.
Simple question answering (QA) bots 
already serve millions of users 
through Amazon's Alexa, Apple's Siri, 
Google's Now, and Microsoft's Cortana.
These bots typically carry out single-exchange conversations, 
but we aspire to develop more general dialogue agents, 
approaching the breadth of capabilities 
exhibited by human interlocutors.
In this work, we consider task-oriented bots~\cite{williams2004characterizing}, 
agents charged with conducting a multi-turn dialogue
to achieve some task-specific goal. 
In our case, we attempt to assist a user to book movie tickets.

For complex dialogue systems, it is often impossible 
to specify a good policy \emph{a priori}
and the dynamics of an environment may change over time.
Thus, learning policies online and interactively 
via reinforcement learning (RL) 
has emerged as a popular approach~\cite{singh2000reinforcement,gavsic2010gaussian,fatemi2016policy}. 
Inspired by RL breakthroughs on Atari and board games~\cite{mnih15human,silver2016mastering},
we employ deep reinforcement learning (DRL) 
to learn policies for dialogue systems.  
Deep Q-network (DQN) agents
typically explore via the $\epsilon$-greedy heuristic,
but when rewards are sparse and action spaces are large 
(as in dialogue systems), this strategy tends to fail.
In our experiments, a randomly exploring Q-learner 
never experiences success in thousands of episodes.

We offer a new, efficient solution 
to improve the exploration of Q-learners. 
We propose a Bayesian exploration strategy that encourages a dialogue agent to explore state-action regions in which the agent is relatively uncertain in action selection.
Our algorithm, the \emph{Bayes-by-Backprop Q-network} (BBQN), 
explores via Thompson sampling,
drawing Monte Carlo samples 
from a Bayesian neural network~\cite{blundell2015weight}.
In order to produce the temporal difference targets for Q-learning,
we must generate predictions 
from a frozen target network~\cite{mnih15human}. 
We show that using the maximum a posteriori (MAP) assignments 
to generate targets results in better performance (in addition to being computationally efficient).
We also demonstrate the effectiveness of \emph{replay buffer spiking} (RBS), a simple technique 
in which we pre-fill the experience replay buffer with a small set of transitions harvested from a na\"{i}ve, 
but occasionally successful, rule-based agent.
This technique proves essential for both BBQNs and standard DQNs. 

We evaluate our dialogue agents on two variants of a movie-booking task.
Our agent interacts with a user to book a movie.
Success is determined at the end of the dialogue if a movie has been booked that satisfies the user. 
We benchmark our algorithm and baselines using an agenda-based user simulator similar to~\namecite{schatzmann2007statistical}. To make the task plausibly challenging, our simulator introduces random mistakes to account for the effects of speech recognition and language understanding errors. 
In the first variant, our environment remains fixed for all rounds of training.
In the second variant, we consider a non-stationary, domain-extension environment.
In this setting, new attributes of films become available over time,
increasing the diversity of dialogue actions available to both the user and the agent.
Our experiments on both the stationary and domain-extension environments demonstrate 
that BBQNs outperform DQNs using
either $\epsilon$-greedy exploration, 
Boltzmann exploration, or the bootstrap approach introduced by~\namecite{osband2016deep}. Furthermore, the real user evaluation results consolidate the effectiveness of our approach that BBQNs are more effective than DQNs in exploration. Besides, we also show that all agents only work given replay buffer spiking, although the number of pre-filled dialogues can be small.

\section{Task-Oriented dialogue systems}
\label{sec:dialogue}
\begin{figure}
\centering
\includegraphics[width=1.0\linewidth]{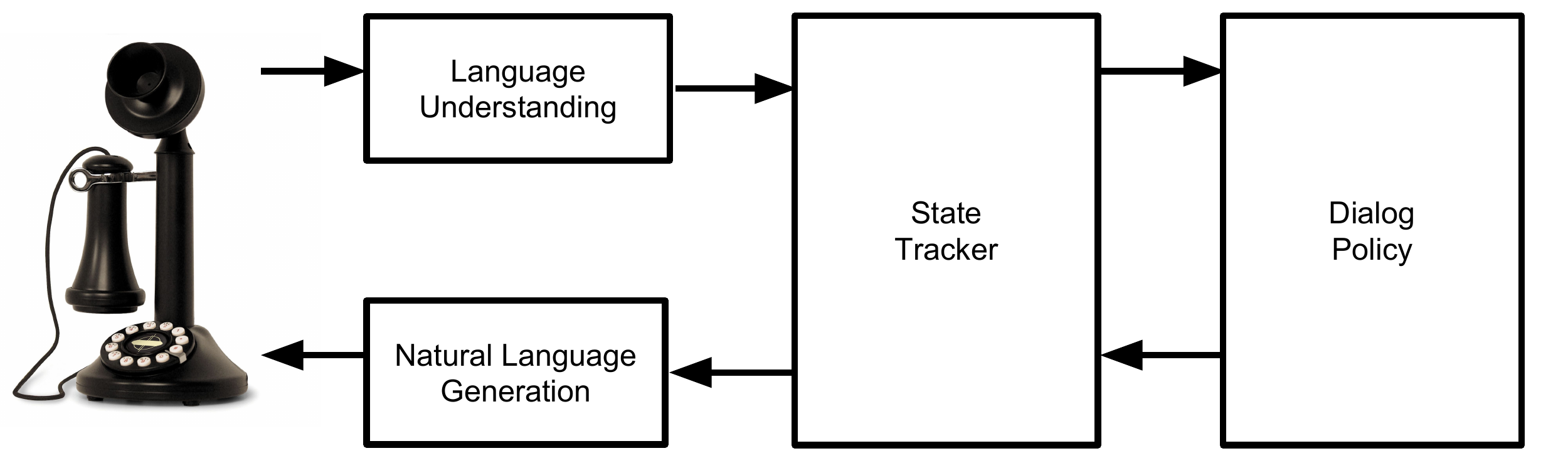}
\caption{Components of a dialogue system}
\label{fig:dialogflow}
\end{figure}

In this paper, we consider goal-oriented dialogue agents, specifically one that aims to help users to book movie tickets.
Over the course of several exchanges, the agent gathers information such as movie name, theater and number of tickets, and ultimately completes a booking. A typical dialogue pipeline is shown in Figure~\ref{fig:dialogflow}. In every turn of a conversation, the \emph{language understanding} module converts raw text into structured semantic representations known as \emph{dialog-acts}, which pass through the \emph{state-tracker} to maintain a record of information accumulated from previous utterances.  The \emph{dialogue policy} then selects an action (to be defined later)  which is transformed to a natural language form by a \emph{generation} module.  The conversation continues until the dialogue terminates.  A numerical reward signal is used to measure the utility of the conversation.  Details of this process are given below.


\paragraph{Dialog-acts} Following~\namecite{schatzmann2007statistical},
we represent utterances as dialog-acts,
consisting of a single \emph{act}
and a (possibly empty) collection of \emph{(slot=value)} pairs,
some of which are \emph{informed} 
while others are \emph{requested} (value omitted).
For example, the utterance, ``I'd like to see \emph{Our Kind of Traitor} tonight in Seattle'' maps to the structured semantic representation \emph{request(ticket, moviename=Our Kind of Traitor, starttime=tonight, city=Seattle)}.

\paragraph{State tracker}
Other than information inferred from previous utterances, the state-tracker may also interact with a database,
providing the policy with information 
such as how many movies match the current constraints.
It then \emph{de-lexicalizes} the dialog-act, allowing the dialogue policy
to act upon more generic states.
The tracked state of the dialogue, consisting of a representation of the conversation history and several database features, is passed on to the policy to select actions.

\paragraph{Actions}
Each action is a de-lexicalized \emph{dialog-act}.
In the movie-booking task, we consider a set of $39$ actions.
These include basic actions such as \emph{greeting(), thanks(), deny(), confirm\_question(), confirm\_answer(), closing()}.
Additionally, we add two actions for each slot: one to inform its value and the other to request it.  
The pipeline then flows back to the user.
Any slots informed by the policy are then filled in by the state tracker. 
This yields a structured representation such as 
\emph{inform(theater=Cinemark Lincoln Square)},
which is then mapped by a natural language generation module to a textual utterance, such as ``This movie is playing tonight at Cinemark Lincoln Square.''

The conversation process above can be naturally mapped to the reinforcement learning (RL) framework, as follows~\cite{levin1997learning}.  The RL agent navigates a Markov decision process (MDP), 
interacting with its environment 
over a sequence of discrete steps~\cite{sutton1998reinforcement}.
At step $t\in\{1,2,\ldots\}$, the agent observes the current state $s_t$, and chooses some action $a_t$ according to a policy $\pi$.
The agent then receives reward $r_{t}$ 
and observes new state $s_{t+1}$, 
continuing the cycle until the episode terminates. 
In this work, we assume that the set of actions, denoted $\mathcal{A}$, is finite.  In our dialogue scenario, the state-tracker produces states, actions are the de-lexicalized dialog-acts described earlier, state transitions are governed by the dynamics of the conversation, and a properly defined reward function is used to measure the degree of success of a dialogue.  In our experiment, for example, success corresponds to a reward of $40$, failure to a reward of $-10$, 
and we apply a per-turn penalty of -1 to encourage pithy exchanges.

The goal of RL is to find an optimal policy to maximize long-term reward. The Q-function measures, for every state-action pair $(s,a)$, the maximum expected cumulative discounted reward achieved by choosing $a$ in $s$ and then following an optimal policy thereafter: $Q^*(s,a) = \max_\pi \mathbbm{E}_\pi\left[\sum_{i=0}^\infty \gamma^i r_{t+i} \mid s_t=s, a_t=a \right]$, where $\gamma\in(0,1)$ is a discount factor.  Owing to large state spaces, most practical reinforcement learners approximate the Q-function by some parameterized model $Q(s,a;\theta)$.  An example, as we used in this paper, is a neural network, where $\theta$ represents the set of weights to be learned.  Once a good estimate of $\theta$ is found so that $Q(\cdot,\cdot;\theta)$ is a good approximation of $Q(\cdot,\cdot)$, the greedy policy, $\pi(s;\theta)=\arg\max_a Q(s,a;\theta)$, is a near-optimal policy~\cite{sutton1998reinforcement}.
A popular way to learn a neural-network-based Q-function is known as DQN~\cite{mnih15human}; see the appendix 
for more details.

\section{Bayes-by-Backprop}
\label{sec:bbb}
Bayes-by-Backprop~\cite{blundell2015weight}
captures uncertainty information from neural networks 
by maintaining a probability distribution 
over the weights in the network.
For simplicity, we explain the idea 
for multilayer perceptrons (MLPs). 
An $L$-layer MLP for model 
$P(\boldsymbol{y}|\boldsymbol{x}, \boldsymbol{w})$ is parameterized by weights $\boldsymbol{w} = \{W_l, b_l\}_{l=1}^{L}$:
$\boldsymbol{\hat{y}}= W_L \cdot \phi(W_{L-1} \cdot ... \cdot \ \phi(W_1 \cdot \boldsymbol{x} + b_1) + ... + b_{L-1}) + b_L\,,$
where $\phi$ is an activation function such as sigmoid, tanh, or rectified linear unit (ReLU). 
In standard neural network training, 
weights are optimized by SGD 
to minimize a loss function such as squared error.  

With Bayes-by-Backprop, we impose a prior distribution 
over the weights, $p(\boldsymbol{w})$, 
and learn the full posterior distribution, $p(\boldsymbol{w}|\mathcal{D}) \propto p(\boldsymbol{w}) p(\mathcal{D}|\boldsymbol{w})$, 
given training data 
$\mathcal{D} = 
\{ \boldsymbol{x}_i,\boldsymbol{y}_i \}_{i=1}^N$.  
In practice, however, computing an arbitrary posterior distribution can be intractable.
So, we instead approximate the posterior 
by a variational distribution, 
$q(\boldsymbol{w}|\theta)$.
In this work, we choose $q$ 
to be a Gaussian with diagonal covariance, i.e., 
each weight $w_i$ is sampled from 
$\mathcal{N}(\mu_i,\sigma_i^2)$.
To ensure that all $\sigma_i$ remain strictly positive,
we parameterize $\sigma_i$  
by the softplus function $\sigma_i = \log (1 + \mbox{exp}(\rho_i))$, 
giving variational parameters $\theta = \{(\mu_i, \rho_i) \}_{i=1}^{D}$ for a $D$-dimensional weight vector $\boldsymbol{w}$. 


We learn these parameters by minimizing \emph{variational free energy}~\cite{hinton1993keeping}, the KL-divergence between the variational approximation $q(\boldsymbol{w}|\theta)$
and the posterior $p(\boldsymbol{w}|\mathcal{D})$:
\begin{eqnarray}
\theta^* &=& \mbox{argmin}_{\theta} \mbox{KL} 
[q(\boldsymbol{w}|\theta) || p(\boldsymbol{w}| \mathcal{D})] \nonumber \\
&=& \mbox{argmin}_{\theta} \Big\{ \mbox{KL}[q(\boldsymbol{w}|\theta) || p(\boldsymbol{w})]
- \mathbbm{E}_{q(\boldsymbol{w}|\theta)}
[\log p(\mathcal{D}|\boldsymbol{w})] \Big\} \nonumber \,.
\label{eqn:var-free-energy}
\end{eqnarray}

When $\boldsymbol{w}$ is sampled from $q$, the above objective function 
can be estimated by its empirical version:
$f(\mathcal{D}, \theta) =
\log q(\boldsymbol{w}|\theta) - \log p(\boldsymbol{w}) - \log p(\mathcal{D}|\boldsymbol{w})$.
It can be minimized by SGVB, using the reparametrization trick popularized by~\namecite{kingma2013auto}.
See appendix 
for more details.



\section{BBQ-networks}
\label{sec:bbq}
We are now ready to introduce \emph{BBQN}, our algorithm for learning dialogue policies with deep learning models.  BBQN builds upon the deep Q-network, or DQN~\cite{mnih15human}, and uses a Bayesian neural network to approximate the Q-function and the uncertainty in its approximation.  Since we work with fixed-length representations of dialogues, we use an MLP, but extending our methodology to recurrent or convolutional neural networks is straightforward. 

\paragraph{Action selection}
A distinct feature of BBQN is that it explicitly quantifies uncertainty in the Q-function estimate, which can be used to guide exploration. In DQN, the Q-function is represented by a network with parameter $\boldsymbol{w}$.  BBQN, in contrast, maintains a distribution $q$ over $\boldsymbol{w}$.  As described in the previous section, $q$ is a multivariante Gaussian with diagonal covariance, parameterized by $\theta=\{(\mu_i,\rho_i)\}_{i=1}^D$.  In other words, a weight $w_i$ has a posterior distribution $q$ that is $\mathcal{N}(\mu_i,\sigma_i^2)$ where $\sigma_i=\log(1+\exp(\rho_i))$.

Given a posterior distribution $q$ over $\boldsymbol{w}$, a natural and effective approach to exploration is posterior sampling, or Thompson Sampling~\cite{thompson33likelihood,chapelle2011empirical,osband2013more}, in which actions are sampled according to the posterior probability that they are optimal in the current state. Formally, given a state $s_t$ and network parameter $\theta_t$ in step $t$, an action $a$ is selected to be $a_t$ with the probability 
$\Pr(a_t = a | s_t, \theta_t) =$
\begin{eqnarray}
\int_{\boldsymbol{w}} \mathbf{1}\{\ Q(s_t,a;\boldsymbol{w}) > Q(s,a';\boldsymbol{w}), \forall a' \ne a\} \cdot dq(\boldsymbol{w}|\theta_t)\,.
\label{eqn:action-posterior}
\end{eqnarray}
Computing these probabilities is usually difficult, but fortunately all we need is a \emph{sample} of an action from the corresponding multinomial distribution. To do so, we first draw $\boldsymbol{w}_t \sim q(\cdot|\theta_t)$, then set $a_t = \argmax_a Q(s_t,a;\boldsymbol{w}_t)$.  It can be verified that this process samples actions with the same probabilities given in the Equation~\ref{eqn:action-posterior}. 
%
We have also considered integrating the $\epsilon$-greedy approach, exploring by Thompson sampling with probability $1-\epsilon$ and uniformly at random with probability $\epsilon$. But empirically, uniform random exploration confers no supplementary benefit for our task.

\paragraph{BBQN}
The BBQN is initialized by a prior distribution $p$ over $\boldsymbol{w}$.  It consists of an isotropic Gaussian whose variance $\sigma_p^2$ is a single hyper-parameter 
introduced by our model.
We initialize the variational parameters to match the prior. So $\boldsymbol{\mu}$ is initialized to the zero vector $\boldsymbol{0}$ 
and the variational standard deviation $\boldsymbol{\sigma}$ matches the prior $\sigma_p$ for each weight.
Note that unlike conventional neural networks, we need not assign the weights randomly because sampling breaks symmetry.
As a consequence of this initialization, from the outset, the agent explores uniformly at random.
Over the course of training, as the experience buffer fills, the mean squared error starts to dominate the objective function and the variational distribution moves further from the prior.

Given experiences of the form $\mathcal{T}=\{(s,a,r,s')\}$ consisting of transitions collected so far, we apply a Q-learning approach to optimize the network parameter, in a way similar to DQN~\cite{mnih15human}.  To do so, we maintain a frozen, but periodically updated, copy of the same BBQN, whose parameter is denoted by $\tilde{\theta}=\{(\tilde{\mu}_i,\tilde{\rho}_i)\}_{i=1}^D$.  For any transition $(s,a,r,s')\in\mathcal{T}$, this network is used to compute a target value $y$ for $Q(s,a;\theta)$, resulting in a regression data set $\mathcal{D}=\{(x,y)\}$, for $x=(s,a)$.  We then apply the Bayes-by-backprop method described in the previous section to optimize $\theta$, until it converges when $\tilde{\theta}$ is replaced by $\theta$.  There are two ways to generate the target value $y$.

The first uses a Monte Carlo sample from the frozen network, $\tilde{\boldsymbol{w}} \sim q(\cdot|\tilde{\theta})$, to compute the target $y$: $y=r+\gamma\max_{a'}Q(s',a';\tilde{\boldsymbol{w}})$.  To speed up training, for each mini-batch, we draw one sample of $\tilde{\boldsymbol{w}}$ for target generation, and one sample of $\boldsymbol{w}$ for sample-based variational inference (see previous section).  With this implementation, the training speeds of BBQN and DQN are roughly equivalent.

The second uses maximum a posterior (MAP) estimate to compute $y$: $y=r+\gamma\max_{a'}Q(s',a';\tilde{\mu})$.  This computationally more efficient choice is motivated by the observation that, since we only require the uncertainty estimates for exploration, it may not be necessary to sample from the frozen network for synthesizing targets.  Furthermore, early in training, the predictive distribution of the networks has high variance, resulting in a large amount of noise in target values that can slow down training.
%

\paragraph{BBQN with intrinsic reward}
Variational Information Maximizing Exploration (VIME)~\cite{houthooft2016vime} introduces an exploration strategy based on maximizing the information gain about the agent's belief of environment dynamics. It adds an \textit{intrinsic reward} bonus to the reward function, which quantifies the agent's \emph{surprise}: $r'(s_t,a_t,s_{t+1}) = r(s_t,a_t) + \eta \mathbb{D}_{KL}[p(\theta|\xi_t,a_t,s_{t+1})||p(\theta|\xi_t)]$, (where $\xi_t$ is defined as the history of the agent up until time step $t$: $\xi_t = \{s_1, a_1, . . . , s_t\}$), and has demonstrated strong empirical performance. We explore a version of BBQNs that incorporates the intrinsic reward from VIME, terming the approach BBQN-VIME-MC/MAP. The BBQN-VIME variations encourage the agents to explore the state-action regions that are relatively unexplored and in which BBQN is relatively uncertain in action selection. In our full-domain experiment, 
both BBQN and BBQN-VIME variations achieve similar performance with no significant difference, 
but in domain-extension experiments, we observe that BBQN-VIME-MC slightly outperforms BBQN-MAP. 

\paragraph{Replay buffer spiking}
In reinforcement learning, there are multiple sources of uncertainty. These include uncertainty over the parameters of our model and uncertainty over unseen parts of the environment.
BBQN addresses parameter uncertainty but it can struggle given extreme reward sparsity.
Researchers use various techniques to accelerate learning in these settings. 
One approach is to leverage prior knowledge, as by reward shaping or imitation learning. Our approach falls into this category. 
Fortunately, in our setting, it's easy to produce a few successful dialogues manually.
Even though the manual dialogues do not follow an optimal policy, 
they contain some successful movie bookings,
so they indicate the existence of the large ($+40$) reward signal. 
Pre-filling the replay buffer with these experiences dramatically improves performance (Figure \ref{fig:warmstart-plots}).
For these experiments, we construct a simple rule-based agent that, while sub-optimal ($18.3\%$ success rate), achieves success sometimes.
In each experiment, we harvest $100$ dialogues of experiences from the rule-based agent, 
adding them to the replay buffer.
We find that, in on our task, RBS is essential for both BBQN and DQN approaches. Interestingly, performance does not strictly improve with the number pre-filled dialogues (Figure \ref{fig:warmstart-plots}). 
Note that replay buffer spiking is different from imitation learning. 
RBS works well with even a small number of warm-start dialogues,
suggesting that it is helpful to communicate 
even the very \emph{existence} of a big reward. 
We find that even one example of a successful dialogue 
in the replay buffer could successfully jump-start a Q-learner.

\section{Experiments}
\label{sec:experiments}
We evaluate our methods on two variants of the movie-booking task. 
In our experiments, we adapt the publicly available\footnote{\url{https://github.com/MiuLab/UserSimulator}} simulator described in~\namecite{li2016user}. In the first, the agent interacts with the user simulator over $400$ rounds.
Each round consists of $50$ simulated dialogues, 
followed by $2$ epochs of training. 
All slots are available starting from the very first episode. 
In the second, 
we test each model's ability to adapt to domain extension by 
periodically introducing new slots.
Each time we add a new slot, 
we augment both the state space and action space.
We start out with only the essential slots: [\textit{date, ticket, city, theater, starttime, moviename, numberofpeople, taskcomplete}]
and train for $40$ training rounds up front.
Then, every $10$ rounds, we introduce a new slot in a fixed order. 
For each added slot, the state space and action space grow accordingly.
This experiment terminates after $200$ rounds.  In both experiments, quantifying uncertainty in the network weights is important to guide effective exploration.

To represent the state of the dialogue at each turn, we construct a $268$ dimensional feature vector, 
consisting of the following:
(i) one-hot representations of the \emph{act} and \emph{slot} corresponding to the current user action, with separate components for requested and informed slots;
(ii) corresponding representations of the \emph{act} and \emph{slot} corresponding to the last agent action;
(iii) a bag of \emph{slots} corresponding to all previously filled slots over the course of the dialog history;
(iv) both a scalar and one-hot representation of the current turn count; 
and (v) counts representing the number of results from the knowledge base that match each presently filled-in constraint (informed slot) as well as the intersection of all filled-in constraints.
For domain-extension experiments, features corresponding to unseen slots take value $0$ until they are seen. 
When domain is extended, we add features and corresponding weights to input layer, initializing the new weights to $0$ 
(or $\mu_i = 0$, $\sigma_i = \sigma_{prior}$ for BBQN), 
a trick due to~\namecite{lipton2015capturing}.


\begin{figure*}[t!]
	\centering
    \begin{subfigure}[b]{0.5\textwidth}
  		\centering
        \includegraphics[width=1.0\columnwidth]{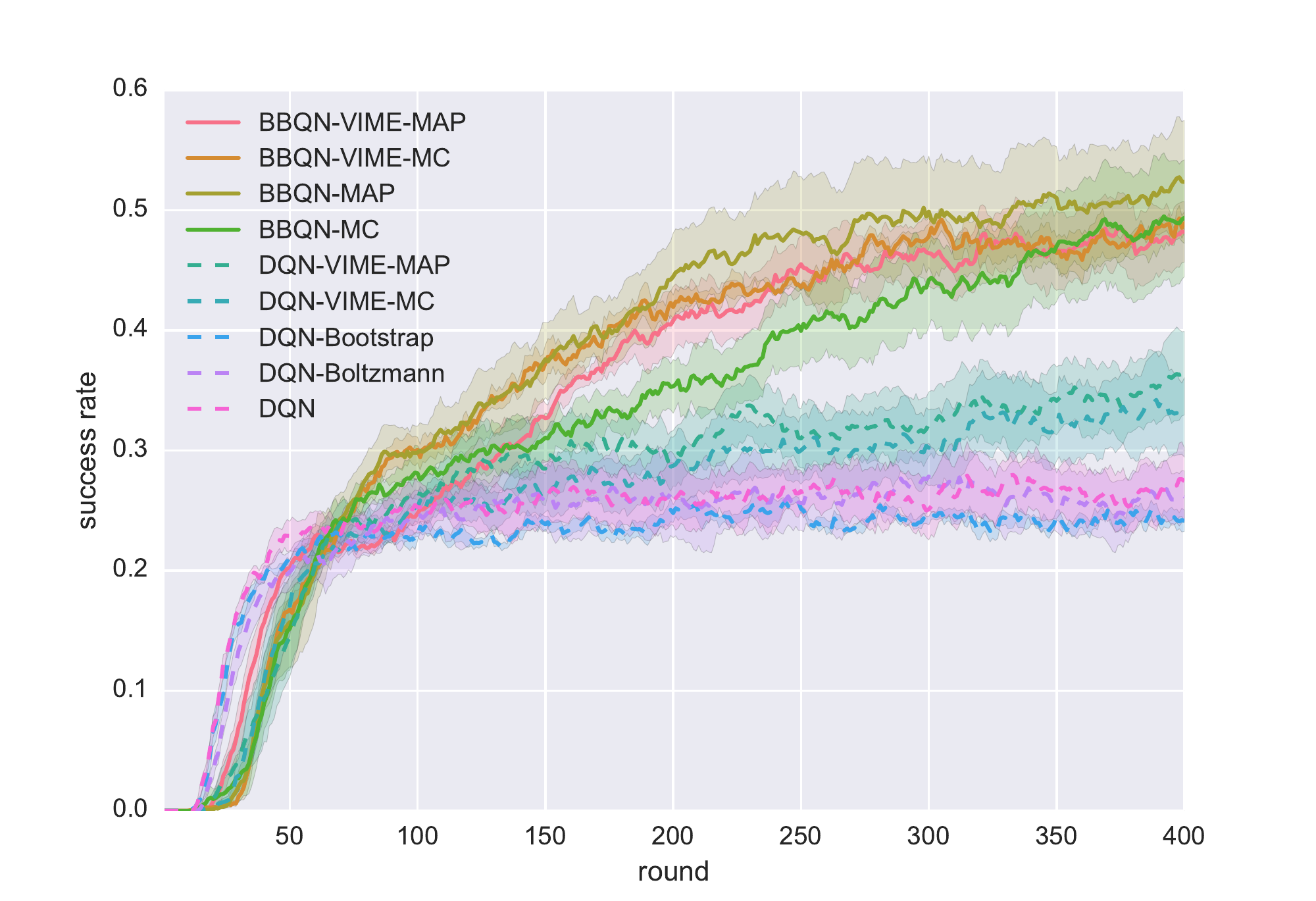}
        \caption{Full domain (success rate)}
\end{subfigure}~
\begin{subfigure}[b]{0.5\textwidth} 
  		\centering
		\includegraphics[width=1.0\columnwidth]{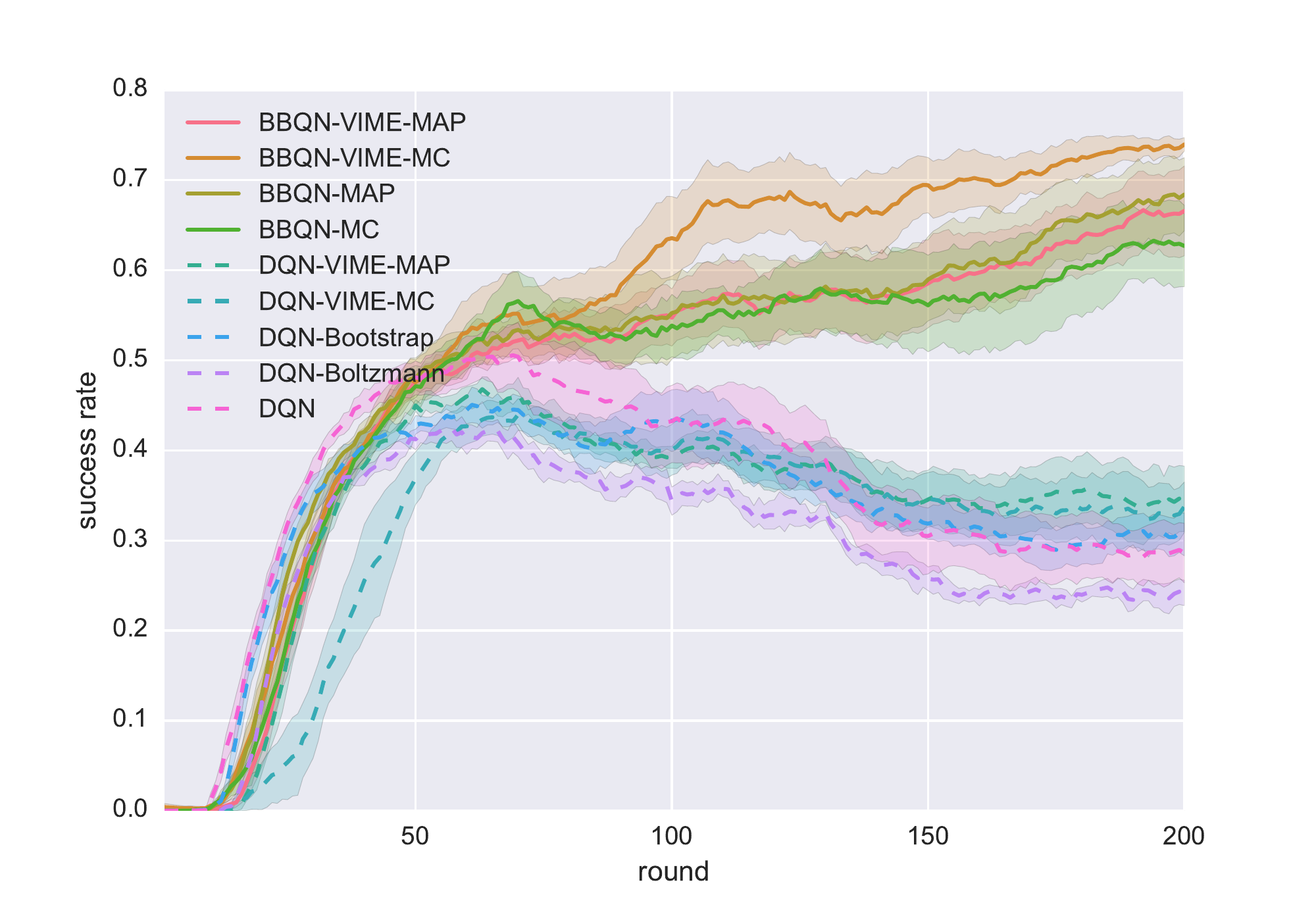}
		\caption{Domain extension (success rate)}
\end{subfigure}
\begin{subfigure}[b]{0.5\textwidth} 
  		\centering
		\includegraphics[width=1.0\columnwidth]{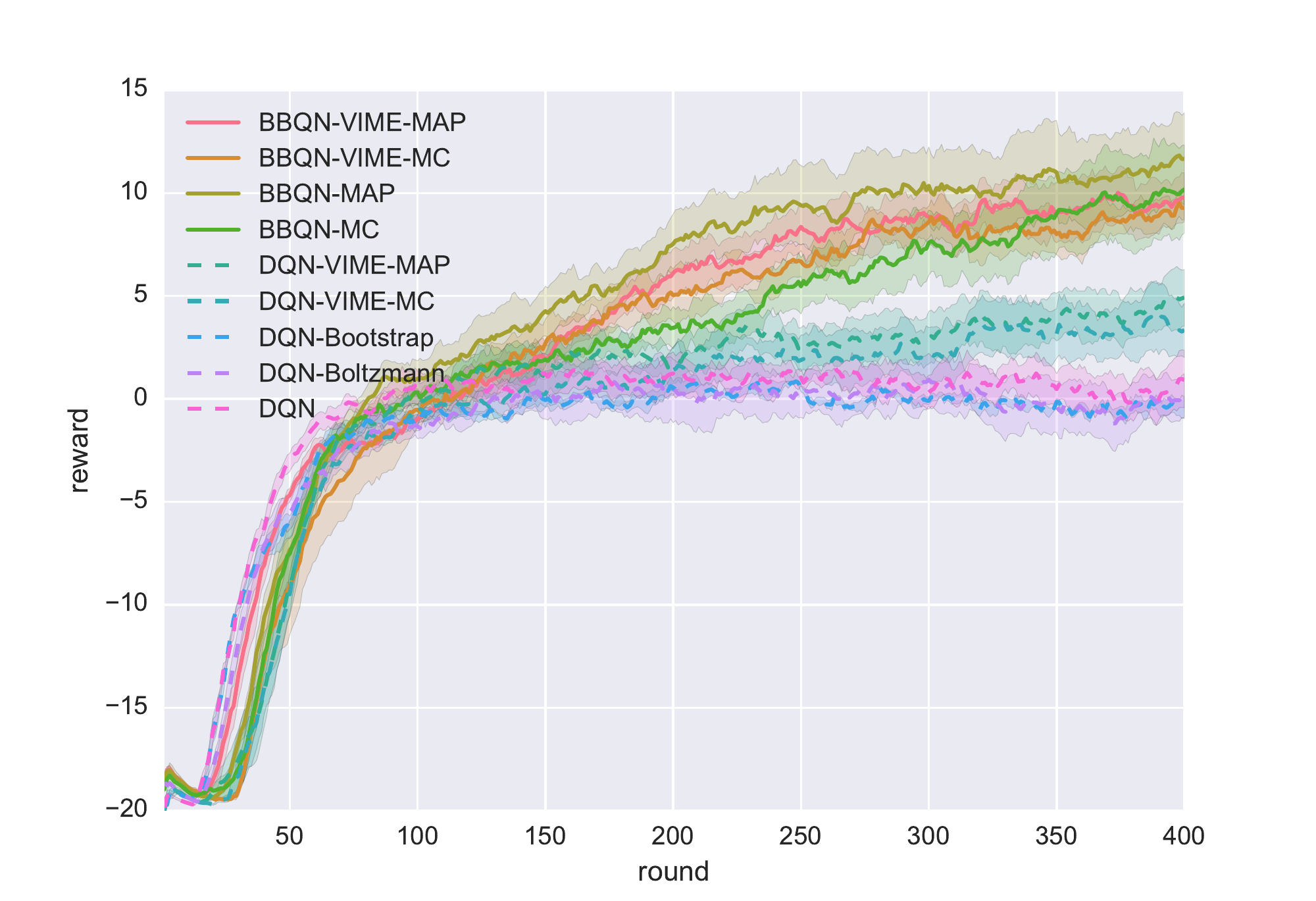}
		\caption{Full domain (reward)}
\end{subfigure}~
\begin{subfigure}[b]{0.5\textwidth} 
  		\centering
		\includegraphics[width=1.0\columnwidth]{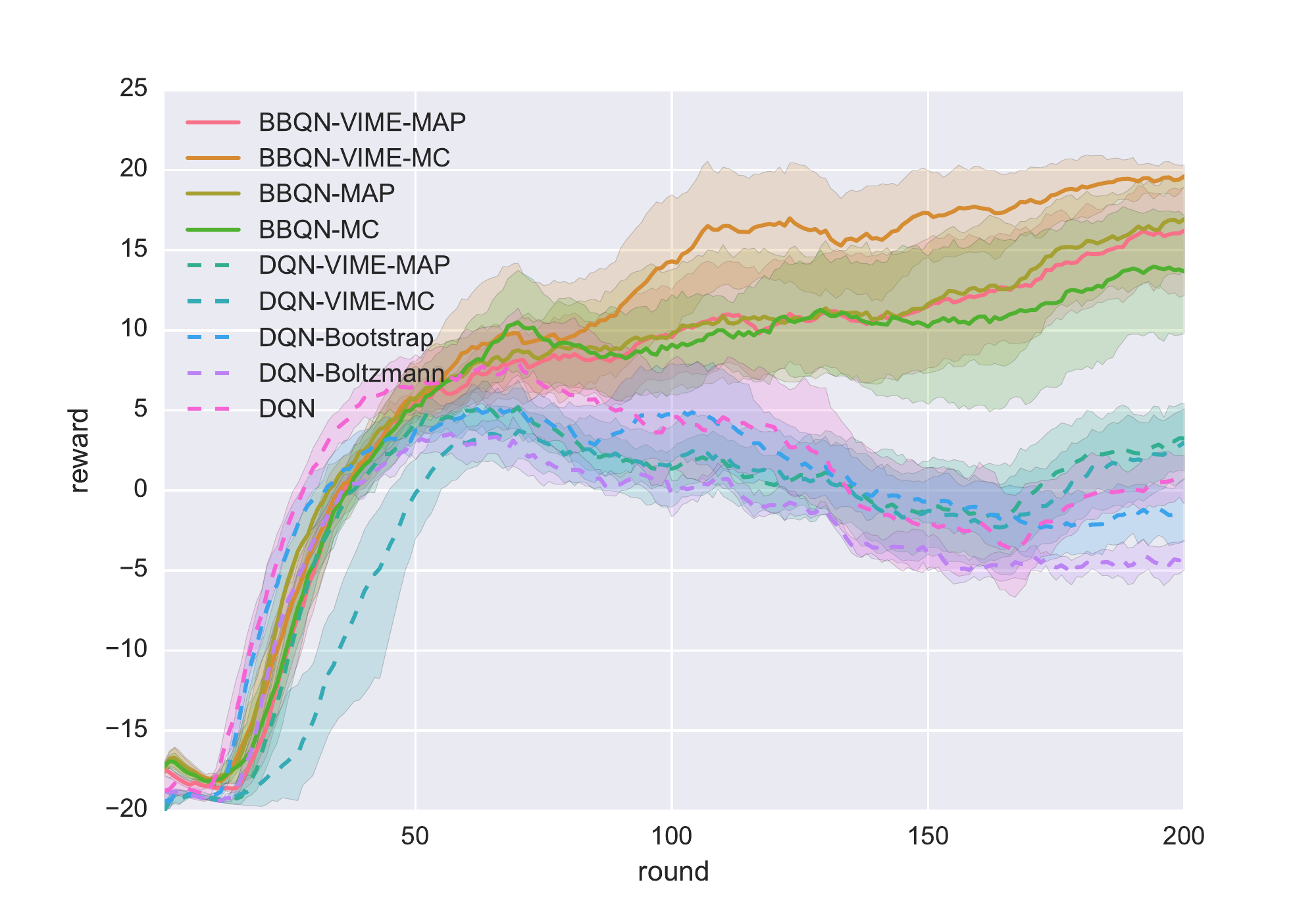}
		\caption{Domain extension (reward)}
\end{subfigure}
\caption{Training plots with confidence intervals for the full domain (all slots available from start) and domain extension problems (slots added every 10 rounds). 
}
\label{fig:training-plots}
\end{figure*}

\begin{table}[t]
\scriptsize
\centering
  \begin{tabular}{ccccc}
    \toprule
    \multirow{2}{*}{Agents} & \multicolumn{2}{c}{Full Domain} & \multicolumn{2}{c}{Domain Extension} \\
\cline{2-3} \cline{4-5}
& {\scriptsize Success Rate} & {\scriptsize Reward} & {\scriptsize Success Rate} & {\scriptsize Reward} \\ 
	\midrule
    {\scriptsize BBQN-VIME-MAP} & 0.4856 & 9.8623 & 0.6813 & 15.8223  \\
    {\scriptsize BBQN-VIME-MC} & 0.4941 & 10.4268 & \textbf{\textcolor{blue}{0.7120}} & \textbf{\textcolor{blue}{17.6261}}  \\
    {\scriptsize BBQN-MAP} & \textbf{\textcolor{blue}{0.5031}} & \textbf{\textcolor{blue}{10.7093}} & 0.6852 & 17.3230 \\
    {\scriptsize BBQN-MC} & 0.4877 & 9.9840 & 0.6722 & 16.1320 \\
    {\scriptsize DQN-VIME-MAP} & 0.3893 & 5.8616 & 0.3751 & 4.9223 \\
    {\scriptsize DQN-VIME-MC} & 0.3700 & 4.9990 & 0.3675 & 4.8270 \\
    {\scriptsize DQN-Bootstrap} & 0.2516 & -0.1300 & 0.3170 & -0.6820 \\
    {\scriptsize DQN-Boltzmann} & 0.2658 & 0.4180 & 0.2435 & -3.4640 \\
    {\scriptsize DQN} & 0.2693 & 0.8660 & 0.3503 & 4.7560 \\
    \bottomrule
  \end{tabular}
  \caption{Final performance of trained agents on 10k simulated dialogues, averaged over 5 runs.}
  \label{tab:final-results}
\end{table}

\paragraph{Training details}
For training, we first use a naive but occasionally successful rule-based agent for RBS. All experiments use $100$ dialogues to spike the replay buffer. We note that experiments showed models to be insensitive to the precise number. After each round of $50$ simulated dialogues, 
the agent freezes the target network parameters $\theta^-$, and then updates the Q- function, training for $2$ epochs, then re-freezes and trains for another $2$ epochs.  
There are two reasons for proceeding in 50-dialog spurts, rather than updating one mini-batch per turn.
First, in a deployed system, 
real-time updates might not be realistic. 
Second, we train for more batches per new turn than is customary in DQN literatures owing to the economic considerations:
computational costs are negligible, 
while failed dialogues either consume human labor (in testing) or confer opportunity costs (in the wild). 

\paragraph{Baseline methods}
To demonstrate the efficacy of BBQN, we compare against $\epsilon$-greedy in a standard DQN. Additionally, we compare against Boltzmann exploration, an approach in which the probability of selecting any action in a given state is determined by a softmax function applied to the predicted Q-values. Here, affinity for exploration is parameterized by the Boltzmann \emph{temperature}.
We also compare to the bootstrapping method of~\namecite{osband2016deep}. For the bootstrap experiments, we use $10$ bootstrap heads, and assign each data point to each head with probability $0.5$. We evaluate all four methods on both the full domain (static) learning problem and on the domain extension problem.

We also tried comparing against Gaussian processes (GP) based approaches.  However, in our setting, due to the high-dimensional inputs and large number of time steps, we were unable to get good results. In our experiments, the computation and memory requirement grow quadratically over time, and memory starts to explode at the 10th (simulation) round.  Limiting data size for GP was not helpful. 
Furthermore, in contrast to \namecite{gavsic2010gaussian} where the state is 3-dimensional, our experiments have 268-dimensional states, making scalability an even bigger challenge.
A recent paper~\cite{fatemi2016policy} compares deep RL (both policy gradient and Q-learning) to GP-SARSA~\cite{engel2005reinforcement} on a simpler dialogue policy learning problem. In order to make Gaussian processes computationally tractable, they rely on sparsification methods~\cite{engel2005reinforcement}, gaining computation efficiency at the expense of accuracy. Despite this undertaking to make GPs feasible and competitive, they found that deep RL approaches outperform GP-SARSA with respect to final performance, regret, and computational expense (by wall-clock). While we consider Gaussian processes to be an evolving area, it is worthwhile to try the Gaussian processes with sparsification methods to compare with deep RL approaches as future work.

\paragraph{Architecture details}
All models are MLPs with ReLU activations. Each network has $2$ hidden layers with $256$ hidden nodes each. We optimize over parameters using Adam~\cite{kingma2014adam} with a batch size of $32$ and initial learning rate of $0.001$, determined by a grid search. To avoid biasing the experiments towards our methods, we determine common hyper-parameters using standard DQN.
Because BBQN confers regularization, we equip DQN models with dropout regularization of $0.5$, shown by~\namecite{blundell2015weight} to confer comparable predictive performance on holdout data.

Each model has additional hyper-parameters. 
For example, $\epsilon$-greedy exploration 
requires an initial value of $\epsilon$ 
and an attenuation schedule.
Boltzmann exploration requires a temperature. 
The bootstrapping-based method of \namecite{osband2016deep} requires both a number of bootstrap heads and the probability that each data point is assigned to each head.
Our BBQN requires that we determine 
the variance of the Gaussian prior distribution 
and the variance of the Gaussian error distribution.

\begin{figure}[!ht]
     \begin{subfigure}[b]{0.5\textwidth}
     \centering
     \includegraphics[width=1.0\columnwidth]{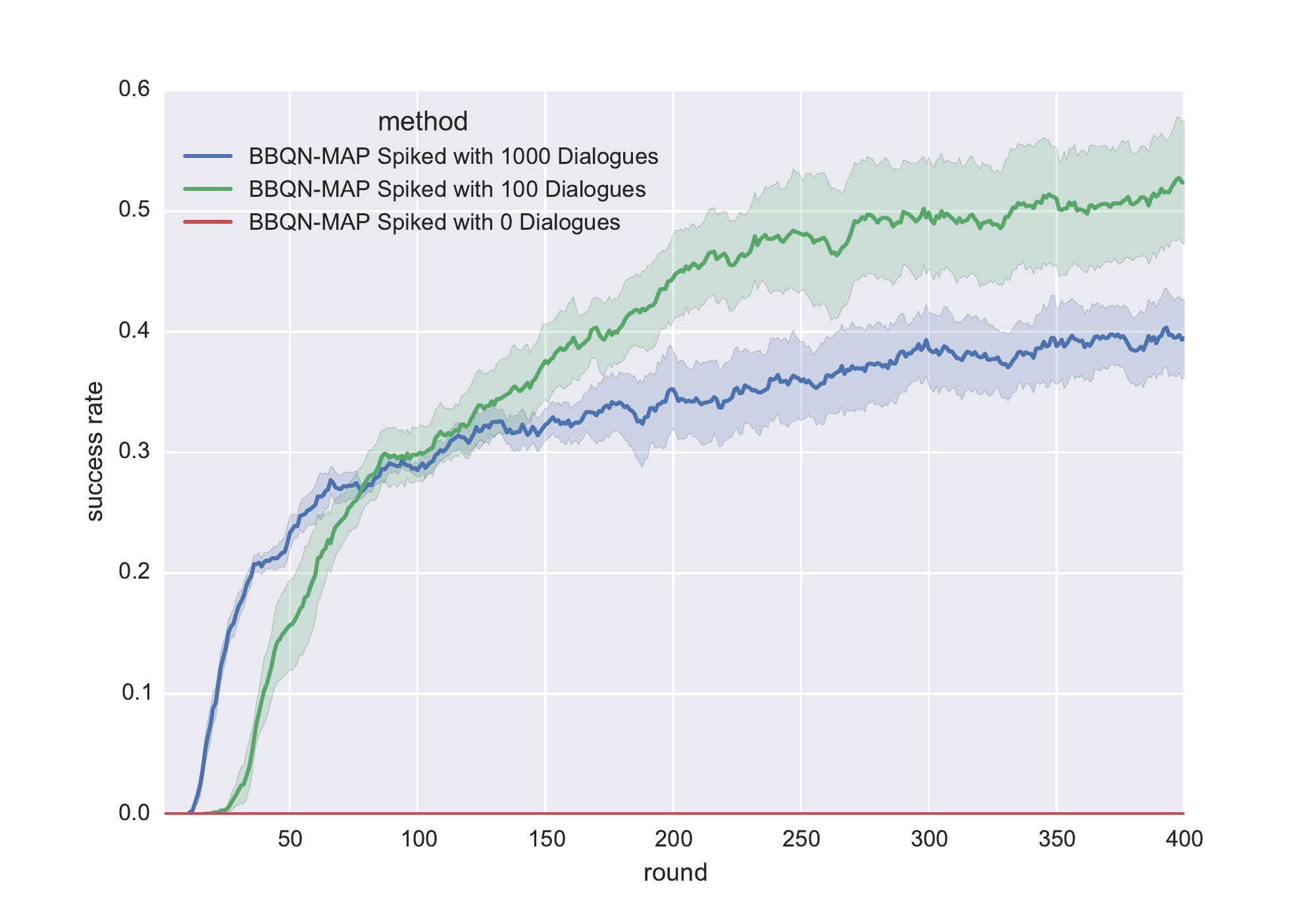}
\end{subfigure}
\begin{subfigure}[b]{0.5\textwidth}
\centering
    \includegraphics[width=1.0\columnwidth]{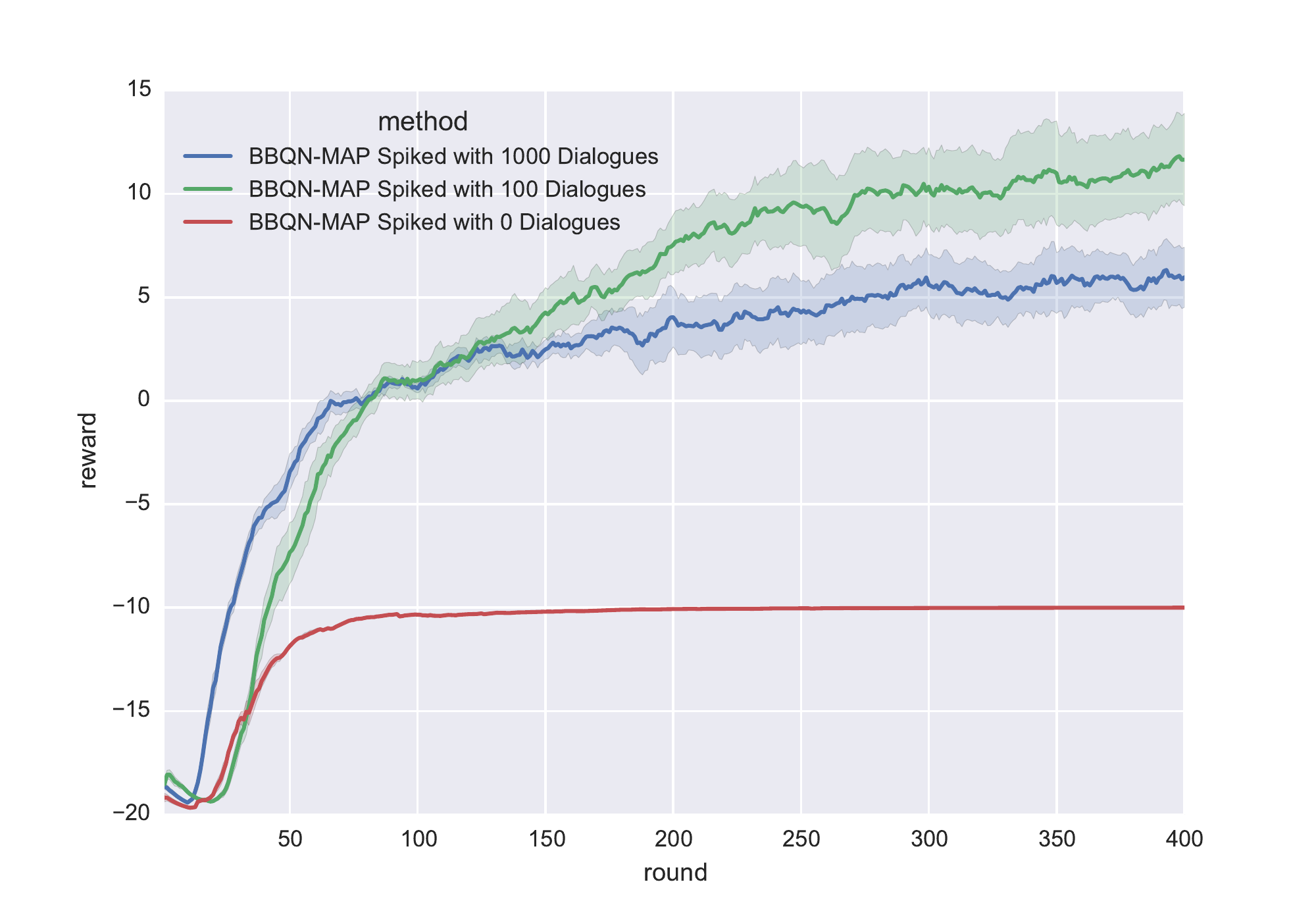}
\end{subfigure}
\caption{
RBS with 100 dialogues improves both success rate (top) and reward (bottom).}
\label{fig:warmstart-plots}
\end{figure}

\paragraph{Simulation results}
As shown in Figure \ref{fig:training-plots}, BBQN variants perform better than the baselines.  In particular, BBQN-MAP performs the best on the full domain setting, BBQN-VIME-MC achieves the best performance on the domain extension setting, with respect to cumulative successes during training 
and final performance of the trained models (Table \ref{tab:final-results}).
Note that the domain extension problem 
becomes more difficult every $10$ epochs, 
so sustained performance corresponds to getting better, while declining performance does not imply 
the policy becomes worse.
On both problems, no method achieves a single success absent RBS.
Evaluating our best algorithm (BBQN-MAP) 
using 0, 100, and 1000 RBS dialogues 
(Figure \ref{fig:warmstart-plots}), 
we find that using $1000$ (as compared to $100$) dialogues, 
our agents learn quickly but that their long-term performance is worse. 
One heuristic to try in the future may be 
to discard pre-filled experiences 
after meeting some performance threshold.

We also considered that perhaps some promising trajectories might never be sampled by the BBQN. Thus, we constructed an experiment exploring via a hybridization of the BBQN's Thompson sampling with the $\epsilon$-greedy approach. With probability $1-\epsilon$, the agent selects an action by Thompson sampling given one Monte Carlo sample from the BBQN and with probability $\epsilon$ the agent selects an action uniformly at random.
However, the uniformly random exploration confers no additional benefit. 

\begin{figure}[!ht]
\begin{subfigure}[b]{0.5\textwidth}
\centering
\includegraphics[width=1.0\columnwidth]{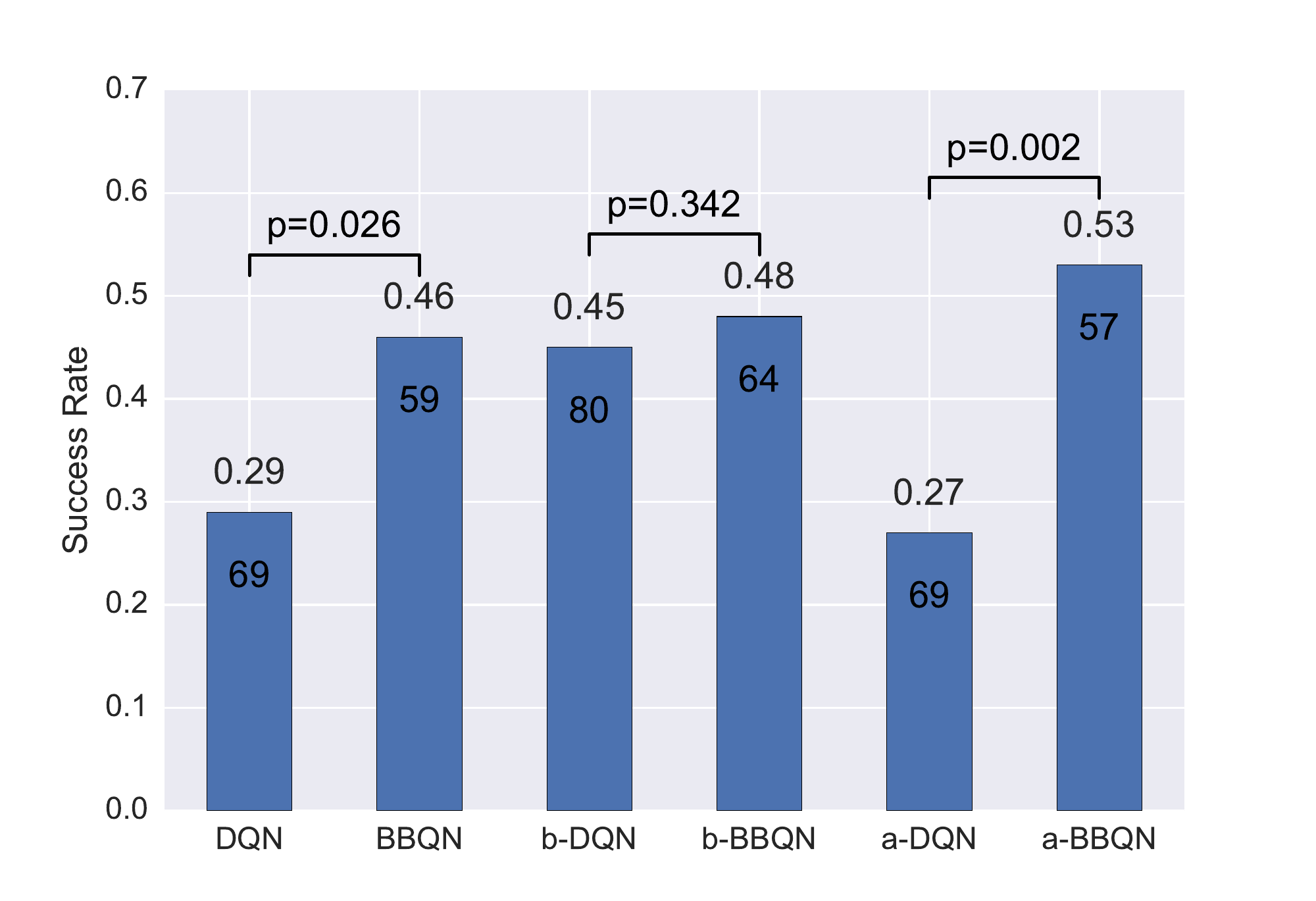}
\caption{Distribution of Success Rate}
\label{fig:realuser_success_plot}
\end{subfigure}
\begin{subfigure}[b]{0.5\textwidth}
\centering
\includegraphics[width=1.0\columnwidth]{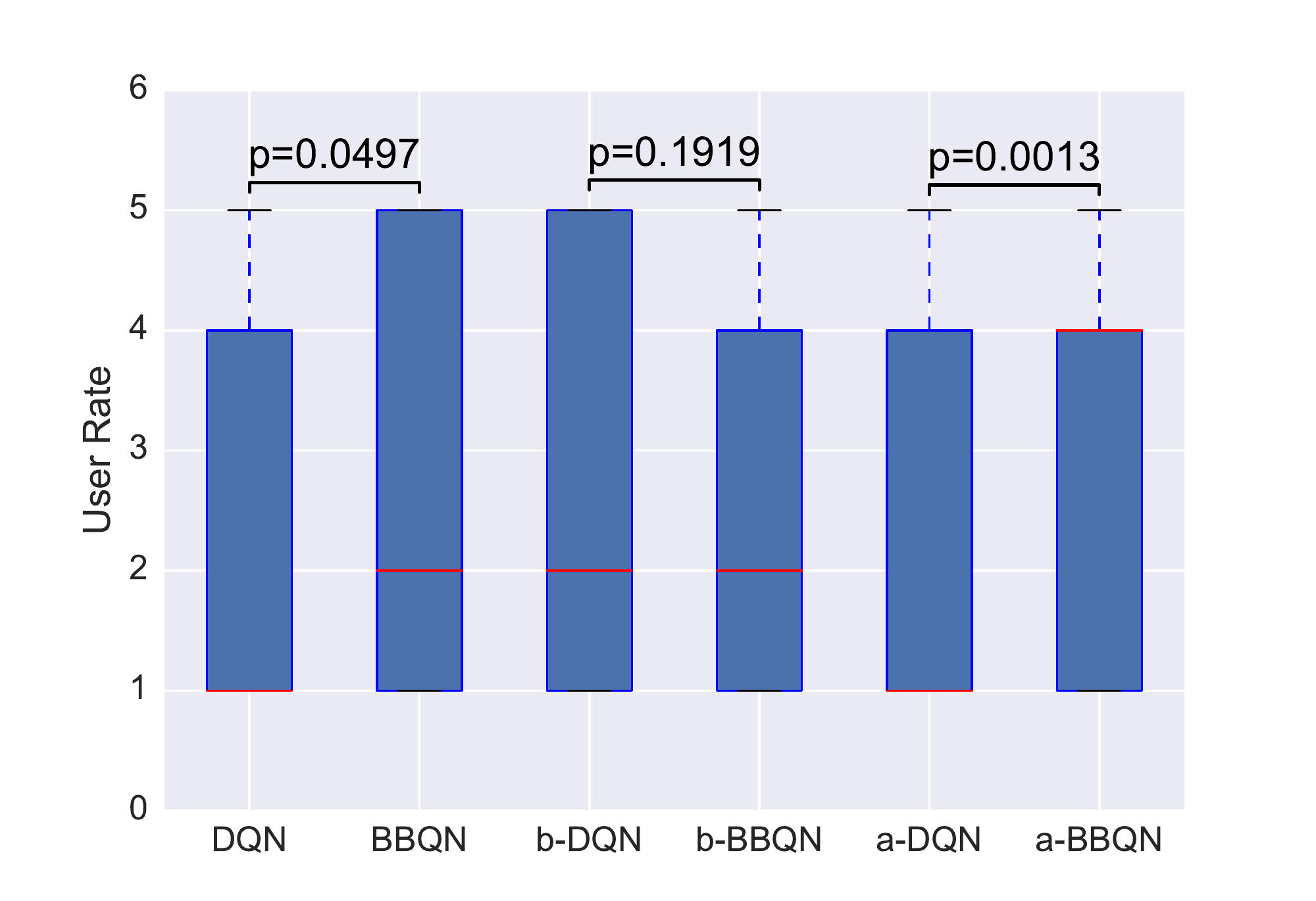}
\caption{Distribution of User Ratings}
\label{fig:realuser_rate_plot}
\end{subfigure}
\caption{Performance of BBQN agent versus DQN agent tested with real users, number of tested dialogues and p-values are indicated on each bar (difference in mean is significant with $p <$ 0.05).}
\label{fig:realuser_study}
\end{figure}

\paragraph{Human evaluation}
We evaluate the agents trained using simulated users against real users, 
recruited from the authors' affiliation.
We conducted the study using the DQN and BBQN-MAP agents. In the full-domain setting, 
the agents were trained with all the slots. 
In the domain-extension setting, 
we first picked DQN (b-DQN) and BBQN (b-BBQN) agents 
before the domain extension at training epoch 40 and the performance of these two agents is tied, nearly 45\% success rate.
From training epoch 40, we started to introduce new slots, and we selected another two agents (a-DQN and a-BBQN) at training epoch 200. 
In total, we compare three agent pairs: 
\{DQN, BBQN\} for full domain, \{b-DQN, b-BBQN\} from before domain extension, and
\{a-DQN, a-BBQN\} from after domain extension. 
In the real user study, 
for each dialogue session, 
we select one of six agents randomly
to converse with a user. 
We present the user with a user goal 
sampled from our corpus.
At the end of each dialogue session, 
the user was asked to give a rating 
on a scale from 1 to 5 based on the naturalness, 
coherence, and task-completion capability of the agent 
(1 is the worst rating, 5 is the best). 
In total, we collected 398 dialogue sessions. Figure~\ref{fig:realuser_success_plot} presents the performance of these agents against real users in terms of success rate. 
Figure~\ref{fig:realuser_rate_plot} shows the comparison in user ratings. In the full-domain setting, the BBQN agent is significantly better than the DQN agent in terms of success rate and user rating. In the domain-extension setting, before domain extension, the performance of both agents (b-DQN and b-BBQN) is tied; after domain extension, the BBQN (a-BBQN) agent significantly outperforms the DQN (a-DQN) in terms of success rate and user rating. 

\section{Related work}
\label{sec:related}
Our paper touches several areas of research, namely Bayesian neural networks, reinforcement learning with deep Q-networks, Thompson Sampling, and dialogue systems.
This work employs Q-learning~\cite{watkins1992q}, a popular method for model-free RL. 
For a broad resource on RL, we point to~\namecite{sutton1998reinforcement}.
Recently, \namecite{mnih15human} achieved 
super-human performance on Atari games using deep Q-learning and incorporating techniques such as experience replay~\cite{lin1992self}.

Efficient exploration remains one of the defining challenges in RL.  
While provably efficient exploration strategies are known for problems with finite states/actions or problems with \emph{nice} structures~\cite{kakade03sample,asmuth2009bayesian,jaksch10near,li11knows,osband2013more}, 
less is known for the general case, especially when general nonlinear function approximation is used.
The first DQN papers relied upon
the $\epsilon$-greedy exploration heuristic~\cite{mnih15human}.
More recently, \namecite{stadie2015incentivizing} and~\citeauthor{houthooft2016vime}~(\citeyear{houthooft2016vime,houthooft2016curiosity}) introduced approaches to encourage exploration by perturbing the reward function.
\namecite{osband2016deep} attempts to mine uncertainty information by training a neural network with multiple output \emph{heads}. Each head is associated with a distinct subset of the data. This works for some Atari games, but does not confer a benefit for us.
\namecite{chapelle2011empirical} empirically examine Thompson sampling, 
one of the oldest exploration heuristics~\cite{thompson33likelihood}, 
for contextual bandits, which is later shown to be effective for solving finite-state MDPs~\cite{strens2000bayesian,osband2013more}. 

We build on the Bayes-by-backprop method of~\namecite{blundell2015weight}, 
employing the reparameterization trick popularized by~\namecite{kingma2013auto},
and following a long history of variational treatments of neural networks~\cite{hinton1993keeping,graves2011practical}.
After we completed this work, \namecite{kirkpatrick2017overcoming} independently investigated parameter uncertainty 
for deep Q-networks to mitigate catastrophic forgetting issues.
\namecite{blundell2015weight} consider Thompson sampling 
for contextual bandits, 
but do not consider the more challenging case of MDPs. 
Our paper also builds on prior work in task-oriented dialogue systems~\cite{williams2004characterizing,gavsic2010gaussian,wen2016network} 
and RL for learning dialogue policies 
~\cite{Levin00Stochastic,singh2000reinforcement,williams2007partially,gavsic2010gaussian,fatemi2016policy}.
Our domain-extension experiments take inspiration from~\namecite{gavsic2014incremental} and our user simulator is modeled on ~\namecite{schatzmann2007statistical}.

\section{Conclusions}
\label{sec:conclusions}
For learning dialogue policies, BBQNs explore with greater efficiency than traditional approaches.
The results are similarly strong for both static and domain extension experiments in simulation and real human evaluation. 
Additionally, we showed that we can benefit from combining BBQ-learning with other, orthogonal approaches to exploration, such as those work by perturbing the reward function to add a bonus for uncovering surprising transitions, i.e., state transitions given low probability by a dynamics model, or previously rarely seen states~\cite{stadie2015incentivizing,houthooft2016vime,houthooft2016curiosity,bellemare2016unifying}.
Our BBQN addresses uncertainty in the Q-value given the current policy, whereas curiosity addresses uncertainty of the dynamics of under-explored parts of the environment. Thus there is a synergistic effect of combining the approaches. On the domain extension task, BBQN-VIME proved especially promising, outperforming all other methods. We see several promising paths for future work. Notably, given the substantial improvements of BBQNs over other exploration strategies, we would like to extend this work to popular deep reinforcement learning benchmark tasks (Atari, etc.) and other domains, like robotics, where the cost of exploration is high, to see if it confers a comparably dramatic improvement.

\bibliography{bbq}
\bibliographystyle{aaai}

\end{document}